# EDS-MEMBED: Multi-sense embeddings based on enhanced distributional semantic structures via a graph walk over word senses[★]

Eniafe Festus Ayetiran[a,*], Petr Sojka[a] and Vít Novotný[a]

[a]*Department of Visual Computing, Faculty of Informatics, Masaryk University, Botanická 68a, Brno, 602 00, Czech Republic*

Abstract

Several language applications often require word semantics as a core part of their processing pipeline either as precise meaning inference or semantic similarity. Multi-sense embeddings (M-SE) can be exploited for this important requirement. M-SE seeks to represent each word by their distinct senses in order to resolve the conflation of meanings of words as used in different contexts. Previous works usually approach this task by training a model on a large corpus and often ignore the effect and usefulness of the semantic relations offered by lexical resources. However, even with large training data, coverage of all possible word senses is still an issue. In addition, a considerable percentage of contextual semantic knowledge are never learned because a huge amount of possible distributional semantic structures are never explored. In this paper, we leverage the rich semantic structures in WordNet using a graph-theoretic walk technique over word senses to enhance the quality of multi-sense embeddings. This algorithm composes enriched texts from the original texts. Furthermore, we derive new distributional semantic similarity measures for M-SE from prior ones. We adapt these measures to word sense disambiguation (WSD) aspect of our experiment. We report evaluation results on 11 benchmark datasets involving WSD and Word Similarity tasks and show that our method for enhancing distributional semantic structures improves embeddings quality on the baselines. Despite the small training data, it achieves state-of-the-art performance on some of the datasets.

## 1. Introduction

Continuous distributed representations of words (word embeddings) are gaining popularity due to their applicability and effectiveness in many areas of natural language processing (NLP). Since the seminal work of Mikolov et al. [38] and subsequent ones [39, 49], word embeddings have been the preferred word semantic representations. The popularity is at least partly influenced by the report of state-of-the-art performances in several areas of its application. The main strength of word embeddings are their ability to capture semantically related terms together in continuous, low-dimensional vector space based on the principle of *"distributional hypothesis"* [23]. This helps to resolve the problem of the so-called *"dimensionality curse"* inherent in the traditional Bag-of-Words (BOW) model. However, word embeddings are still prone to the problem of conflation of word meanings (polysemy). This is because the representation is based on local co-occurrence of words in the training corpus without regard to their individual meanings. Multi-sense embeddings (M-SE) are an active research approach to tackle the problem of polysemy, with the common goal of producing better representations of texts based on the specific meanings of words as used in particular contexts.

Early studies open a discuss about whether M-SE can improve the quality of text representation and consequently, their performance on tasks in which they are applied. Some works which specifically try to answer this question include Li and Jurafsky [32] and Dubossarsky et al. [15]. While the former reported improvements in some tasks, the latter reported that ground truth polysemy worsens performance on the Word Similarity task. However, several other experimental works [25, 19, 46, 50] reported improvements in performance on tasks in which they were applied.

It is not surprising that different studies have divergent views about the effectiveness of M-SE. This is due to the fact that its success or failure depends largely on two important factors. The first is how representative the training data is in terms of the distributional semantic structure of its content. The second is how accurate the disambiguated tokens are, since sufficient manually sense-annotated data is unavailable. M-SE, like word embeddings are usually evaluated on word-based tasks. However, we agree with one of the submissions of a previous study [15] which posits that a more appropriate task other than Word Similarity should also be explored in the evaluation of M-SE apart from the traditional Word Similarity task. Following this agreement, while we do not rule out the suitability of word-based evaluation, our position is that tasks which require sense-specific inference should also be explored. The intuition behind our belief is that multi-sense embeddings should result in considerable gain in inferring word senses since it distinctly represent each word by their specific senses.

M-SE have been studied using a multi-prototype approach [25, 46] and an approach based on a predefined sense inventory [28, 59]. Our approach is based on the latter and we use WordNet as our sense inventory. One advantage of using a predefined sense inventory is that all possible word senses are considered. This consequently allows for fair evaluation on sense-based tasks.

In order to obtain high-quality embeddings, models are usually trained on large amounts of data. However, even with very large data of several billions of tokens, representative distributional semantic structures are still an issue. The issue is worse for multi-sense embeddings because of the sense-specific representations of words for model training. M-SE is best positioned to tackle this bottleneck of distributional semantics learning because of the sense-specific annotation of words in its development pipeline which offers the possibility to explore semantic relations.

Inspired by the efficacy of the application of lexical semantic relations in algorithms for word sense disambiguation (WSD), we investigate the effect of distributed composition of words' meanings using their senses and that of their semantic relations in WordNet [18]. In reality, words in a piece of text can usually be replaced with semantically related words, with the results still maintaining a set of meaningful piece of texts. For instance, the word *"married"* (past tense of first verb sense of the word *"marry"* in WordNet) in the following statement: *"He got married three years ago"* can easily be replaced with the word *"divorced"* (past tense of second verb sense of the word *"divorce"* in WordNet) resulting in the following meaningful statement *"He divorced three years ago"*. The semantic relations between the two example words (*"marry"* and *"divorce)* in WordNet can either be that of antonymy or entailment. Antonymy is the opposite relationship while entailment can be described as an event or action which is required to occur before the latter can take place. In the same vein, the word *"years"* (the plural of second noun sense of the word *"year"* in WordNet) can be replaced with any of the words *"months"* or *"weeks"* in the two previous statements representing the hyponymy relation. Ayetiran and Agbele [5] and Ayetiran [4] posit that words are best characterized by their specific definitions and explore this assertion for WSD and text classification. Following up on this position, lexically related word senses have something in common in their definitions. For instance, in the latter example, *"year"* and *"month"* have the word *"calendar"* in common. We can in principle link these related words to each of their definitions.

The crux of our hypothesis is that enriching text through linking of semantic relations can significantly enhance learning of corpus-based distributional semantics. It is worthy of mention that our graph-theoretic algorithm is different from the work of Pilehvar and Collier [50] which takes pre-trained word or sense embeddings as input and uses a graph-based technique for extracting semantic biasing words from their synsets in WordNet. In contrast, our approach is a graph walk language generation algorithm which offers the advantage of debiasing in the resulting sense embeddings. This bias is one of the drawbacks of word embeddings since it allows the dominant words irrespective of the meaning to be captured along with more words with which they co-occur. Our goal is to use the semantic relations in WordNet to enhance semantic structures in training data and consequently learning of distributional semantics therein.

In specific terms, the contributions of this work are as follow:

- We develop a graph-theoretic method which leverages WordNet's relational graph structure to enhance learning of distributional semantics

- We derive sense-specific distributional semantic similarity measures from prior measures and adapt it to a new knowledge-based WSD algorithm

In principle, these contributions offer the following advantages:

- Using a lexical database ensures coverage of all possible word senses compared to when only a text corpus is used which is prone to coverage issue irrespective of the size of the corpus

- Based on the evaluation results, training data enhancement with lexical semantic relations improves performance when compared to training without enhancement.

- The small size of the training data offers computational efficiency over state-of-the-art models. This is because the training time is directly proportional to the size of the data

Following our earlier position that the effect of M-SE should also be explored on sense inference tasks, we first evaluate our model using six WSD benchmark datasets from the SemEval workshop series. Additionally, to have an outlook of how our approach performs in comparison with previous studies, a second evaluation was carried out using 5 word similarity benchmarks. The model built from training data without enhanced distributional semantic structure serves as the primary baseline. In all but one of the datasets, the results show that our method improves the primary baselines hence proves our hypothesis. On the WSD experiment, the embeddings-based methods are the secondary baselines. We also compare our results with other methods which cut across diverse approaches. On the Word Similarity task, we make a general comparison since all systems are based on embeddings.

The rest of the paper is structured as follows: Section 2 discusses the related works. In Section 3, we present the graph walk algorithm. The model development is discussed in Section 4. Section 5 discusses the experiments. Discussion of the results is presented in Section 6 while Section 7 concludes the paper.

## 2. Related work

The amount of work on multi-sense embeddings is relatively minimal when compared to word embeddings. One of the earliest attempts to solve the problem of polysemy in word embeddings is the work of Reisinger and Mooney [55]. Each concept is represented as an abstract prototype instance and the occurrences of these instances for all words are clustered. The similarity between two words is then defined as a function of their cluster centroids rather than as a centroid of all words' occurrences. In another work, Huang et al. [25] introduced global context along with the local context into their model architecture. They built multiple vectors per word type by sampling different senses of words using WordNet. They evaluated their work using WordSim-353 dataset [19] and another dataset they created to take context into account. They reported superior performance over baselines.

Neelakantan et al. [46] employed a non-parametric technique to jointly perform WSD and embeddings and then estimate the number of senses per word type. They used a modified version of the skip-gram architecture proposed by [38], which they called the Multiple-sense Skip-gram (MSSG). It also has a non-parametric counterpart called the *Non-parametric Multiple-sense Skip-gram (*NP-MSSG*)*. The core contributions of their work include improved performance and training efficiency (27× faster) over Huang et al. [25].

For mutual benefit of word sense embeddings and disambiguation, Chen et al. [13] proposed a unified approach. Their approach is premised on two ideas: (i) that a quality representation of word senses should capture rich information about words and senses, which in turn should be helpful for WSD and (ii) that WSD can provide well-grounded disambiguated corpus for quality sense representations. WordNet was used as the lexical resource and evaluation was done on Word Similarity and WSD.

Works on usage of pre-defined sense inventory for multi-sense embeddings include [28]. It adopts BabelNet [45] as sense inventory. BabelNet is an integration of WordNet, Wikipedia and other lexical resources. They generated a sense-annotated corpus used for training their model by disambiguating the English Wikipedia dump using Word2Vec [38]. Evaluation of their work was carried out using Word and Relational Similarity tasks. In order to mitigate the effect of bias of frequent senses over infrequent ones, Chen et al. [12] used convolutional neural networks to initialize word embeddings through sentence-level embeddings in WordNet glosses. Evaluation on Word Similarity and analogical reasoning tasks showed improvements in approximately 50% of the test datasets.

AutoExtend [57] like in the work of Pilehvar and Collier [50], takes word embeddings as input and extends them to synset and lexeme embeddings. They coined a name for the premises on which their model is based, called *"synset constraints"*. The components of *"synset constraints"* include: (i) words are sums of their lexemes and (ii) synsets are sums of their lexemes. They used autoencoding architecture which consist of two parts; a part for encoding words as synsets and lexemes and another part for decoding the words. They reported state-of-the-art results on Word Similarity and WSD tasks at the time of reporting. Their primary sense inventory is WordNet but they later extended their work [58] with the use of additional lexical resources which include Freebase [8] and GermaNet [22].

In a slight departure from the previous works, Camacho-Collados et al. [10] presented three multilingual sense representations from lexical knowledge and corpus statistics, classified into lexical, embedded and unified vectors. Their multilingual vectors were built through the use of BabelNet [45]. They evaluated their work using benchmarks from Word Similarity, Sense Clustering, Domain Labeling and Word Sense Disambiguation.

In another work, Nguyen et al. [47] applied a mixture of topic models to multi-sense embeddings using weight induction of different senses of a word. Experimentation and evaluation of this method was done on Word Similarity and Analogy tasks. Rather than a full multi-sense embeddings, Mancini et al. [37] chose a unified joint embeddings of word and senses in the same semantic space using what they called *shallow word-sense connectivity algorithm*. They evaluated this approach using Word Similarity and Sense Clustering.

Athiwaratkun et al. [3] applied probabilistic fastText to multi-sense embeddings. The major component of their work is the ability to capture the senses of sub-word structures, rare, misspelt and unseen words, being an adaptation of original fastText algorithm [7].

More recently, Ruas et al. [59] used a new WSD algorithm, called the *Most Suitable Sense Annotation* for multi-sense embeddings. The efficacy of this algorithm was tested on five Word Similarity benchmarks. Iacobacci and Navigli [27], following the work of Mancini et al. [37], also chose to represent words and senses jointly for the purpose of evaluation on word-to-sense tasks. However, they also carried out evaluation on other word-based tasks such as Synonym Recognition, Outlier Detection and Word Similarity. Their embeddings was developed using the long short-term memory (LSTM) architecture with pre-trained word embeddings as the objective function. Their sense inventory of choice is BabelNet.

In the latest work of Scarlini et al. [61], they built contextualized multi-sense embeddings specifically for multilingual WSD using BabelNet as the sense inventory. The primary source of knowledge for their model are BabelNet and Wikipedia. Similar to the work of Pilehvar and Collier [50] and Rothe and Schütze [57], their multi-sense embeddings algorithm takes word embeddings as input, consequent upon context retrieval from Wikipedia that can characterize a given synset. They carried out evaluation on the noun subsets of 5 WSD benchmark datasets from the SemEval workshop series. Results show some improvements over previous works on the evaluated subsets.

## 3. Graph formulation and walk for distributed sense-annotated text generation

Given a sense-tagged sentence or statement $T$ in a corpus $D$, the graph network $G = (V, E)$ is a graph of vertices $V$ and their connection through a set of edges $E$. Each vertex $v'_i \in V$ represents the original input word sense in a text. Each text is a word sense in WordNet with their associated definitions. The root vertex is a target word sense while other vertices are senses of the tokens in its definition. The root

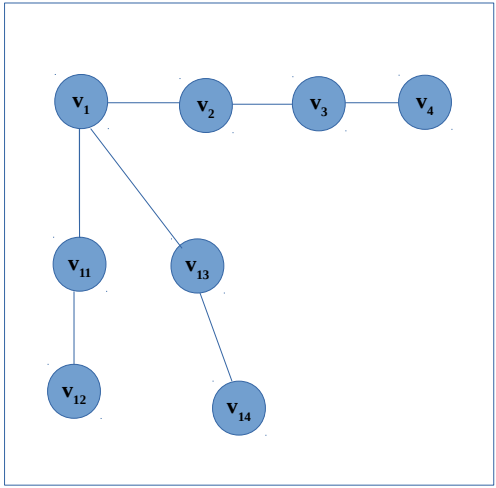

**Figure 1:** Graph $G$ for sense-annotated text generation

**Algorithm 1** Graph $G$ for text generation
---
**Require**: Lexical database $D$
**Input**: Sense-tagged text, $S_I = s_1, s_2, \ldots, s_m \in D$
**Output**: A sequence of sense-tagged text $\{S_O = s_1, s_2, \ldots, s_n\}$

1: **procedure** GRAPHFORMULATE($G(V, E)$)
2:     **for** $s_i \in S_I$ **do**
3:         Select $s_{ij} \leftarrow$ relations for $s_i$
4:         **if** $s_{1j} \neq \phi$ **then**
5:             Construct graph $G, \ni$ vertices,
$$V = \{v_1, v_{1j}, \ldots, v_n\} = \{s_1, s_{1j}, \ldots, s_n\}$$
6:         **else**
7:             Construct graph $G, \ni$ vertices,
$$V = \{v_1, v_1, \ldots, v_n\} = \{s_1, \ldots, s_n\}$$
8:         **end if**
9:     **end for**
10: **end procedure**
11: **procedure** GRAPHWALK($S_O \leftarrow (v_i, v_{ij}, \ldots, v_n)$)
12:     initialize sequence, $S_O = \{\}$
13:     **for** $v_i, v_{ij} \in V$ **do**
14:         Traverse vertices $v_i, v_{ij} \in V$ using depth-first search
15:         Concatenate each traversed vertex to $s_O$
16:     **end for**
17:     return $S_O$
18: **end procedure**

vertex comprises sub-vertices $v_{ij}$ (if any) representing its semantic relations in WordNet. Vertices of words (senses) in the sentence $T$ are connected in a linear sequential order while the sub-vertices are connected to the root vertex in a depth-wise order.

To generate a sequence of enriched sense-tagged text, an empty sequence is first initialized. A depth-first walk is then performed on the graph starting from the root to the destination vertex. At each transition step, a traversed vertex is added to the initialized sequence until all traversed vertices have been added to the sequence. The size of the sequence for each text is given by $\sum v_i v_{ij}$.

An illustration of the graph formulation and walk is given in Figure 1 showing a graph consisting of four vertices (denoted $v_1, v_2, v_3$ and $v_4$), where $v_1$ is the root of the graph, for a sentence $T = (v_1, v_2, v_3, v_4)$. The root represents the sense of a target word under consideration while the other vertices represent its sense-tagged definition in WordNet. The root has four sub-vertices, representing its lexical semantic relations in WordNet. The graph walk results to the following sense-annotated sequence: $(v_1, v_{11}, v_{12}, v_{13}, v_{14}, v_2, v_3, v_4)$.

The summary of the graph formulation and walk procedure is given in Algorithm 1.

## 4. Model development

This section describes the phases involved in the development of the embeddings.

### 4.1. Building the enhanced distributed sense-annotated corpus

One of the most important components in multi-sense embeddings is the input sense-tagged corpus for training a model because the size of the training data and how accurate the sense tags therein will determine how representative the model will be. As a result of inadequate publicly available sense-tagged corpora, annotation of text corpus through WSD has been the tradition. However, a major and open problem in word sense disambiguation is the accuracy. The sense-annotated data used in this work are partially annotated and comprise the following:

- **Semcor corpus**: Semcor [40] is a subset of the Brown corpus that contains about 362 texts comprising over 200,000 words. Some of the texts were fully manually tagged while some have only the verb part-of-speech tagged. It was originally annotated with WordNet 1.6 senses but later mapped to version 3.0.

- **WordNet gloss tags (WNGT)**: WNGT [30] is a manual and automatic annotation of the glosses of the WordNet senses. The choice of WNGT is necessitated by the need to cover all possible word senses.

We used both corpora in the version present in the UFSAC unified corpus Vial et al. [64]. The statistics of the two corpora in the unified corpus are presented in Table 1.

A considerable percentage of the unannotated words are stopwords and punctuations which are not taggable (do not have senses) and therefore excluded from the representation learning. The unannotated words in the two corpora were disambiguated and annotated appropriately using a two-stage method.

In the first stage, an optimized knowledge-based WSD algorithm [5] is adopted for disambiguation. The disambiguation of words in these corpora is relatively easier when compared to tagging a fully unannotated corpora because most words have been manually annotated. To obtain the distribution of contextual information, a slight modification is made to the original disambiguation algorithm. For already annotated words, only the lexical information of the context content words in WordNet were utilized. In the case of unannotated words, the lexical information of all available synsets in WordNet were utilized. All knowledge information are strictly from WordNet. The similarity for the senses of each context word are first computed separately, then a summation of these individual similarities is then computed to obtain the cumulative similarity score. The sense of the target word with the highest cumulative similarity score is the winning sense.

**Table 1**
Statistics of Semcor and WNGT in UFSAC.

| Corpora | Sentences | Words | | Annotated Part of Speech | | | |
|---|---|---|---|---|---|---|---|
| | | Total | Annotated | Nouns | Verbs | Adj. | Adv. |
| Semcor | 37,176 | 778,587 | 229,517 | 87,581 | 89,037 | 33,751 | 19,148 |
| WNGT | 117,659 | 1,634,691 | 496,776 | 232,319 | 62,211 | 84,233 | 19,445 |

In the second stage, a Word2Vec model is trained on the annotated corpora derived from the first stage. The embeddings-based WSD algorithm described in Section 5.3 (using the concept of maximum similarity) is then applied on the untagged words in the original corpora to derive a new tagged corpora with improved accuracy.

The final phase in the distributed corpus generation is the application of the graph walk algorithm to generate sequences of tagged text for each word sense in WNGT. The considered semantic relations used for enriching the text structures include hypernymy, hyponymy, topic domain, antonymy, member meronyms, part meronyms, substance meronyms, member holonyms, part holonyms, substance holonyms, causes, entailment, pertainymy and derivationally related forms. The resulting corpus generated from the graph-based algorithm is then augmented with the fully annotated Semcor corpus. Due to the fact that WordNet synonyms have different sense identifiers, each tagged word sense in the corpus is represented by its offset in WordNet (normalized to 8 digits) which is a unique identifier for all synonyms. Each normalized offset is then appended with its part-of-speech. For example, one of the noun senses of the word *"bank"* is represented as *"08420278n"*, where the digits are the normalized offset in WordNet and the "n" character is the part-of-speech notation denoting that the sense is a noun. Characters "v" and "r" denotes verb and adverb respectively. An adjective can be denoted by either "a" or "s" depending on whether it is a normal adjective or an adjectival satellite. WordNet was used through the NLTK [33].

We give a specific example to further explain the distinction between multi-sense embeddings and traditional word embeddings and how the graph walk algorithm generates text to create the enhanced distributed sense-annotated corpus. In traditional word embeddings, to learn the distributional semantics to represent the target word *"permit"*, the Word2Vec model or any distributed word representation algorithm creates a single representation for the word *"permit"* using the contextual information in its left and right anywhere it appears in the training corpus depending on the window size. This ignores the various senses of the word and embeds the learned distributional semantics into a single representation. For multi-sense embeddings, this approach is different. *"permit"* has a total of six senses in WordNet including three noun and three verb senses.

The idea of multi-sense embeddings is to have a distinct representation for each of these senses. To represent the first verb sense *"Synset('permit.v.01')"* which is defined as *"consent to, give permission"* in that sequential order. The sense tags for each of the words in the definition are utilized if they are already tagged, otherwise they will be tagged through our WSD algorithm. Removing the stopword *"to"*, the words *"consent"*, *"give"* and *"permission"* can be correctly sense-tagged as *"Synset('accept.v.03')"*, *"Synset('give.v.09')"* and *"Synset('permission.n.01')"* respectively. According to the WordNet relational graph, *"Synset('accept.v.03')"* is a hypernym of *"Synset('permit.v.01')"*. The following are its hyponyms: *"Synset('admit.v.02'), Synset('admit.v.03'), Synset('allow.v.10'), Synset('authorize.v.01'), Synset('digest.v.03'), Synset('furlough.v.02'), Synset('give.v.40'), Synset('legalize.v.01'), Synset('privilege.v.01') and Synset('trust.v.02')"*. Based on the depth-first mode of operation of the graph walk algorithm, the distributed text is generated for the sense *"Synset('permit.v.01')"* starting from itself (the root - the target sense) through its sub-vertices (its hypernyms and hyponyms) and then through the sense tags in its definition. The output of the generated sense-tagged text is: *"Synset('permit.v.01') Synset('accept.v.03') Synset('admit.v.02'), Synset('admit.v.03'), Synset('allow.v.10'), Synset('authorize.v.01'), Synset('digest.v.03'), Synset('furlough.v.02'), Synset('give.v.40'), Synset('legalize.v.01'), Synset('privilege.v.01') Synset('trust.v.02') Synset('accept.v.03') Synset('give.v.09') Synset('permission.n.01')"*. The senses are finally represented by their unique offset to give the following output sequence: *"00802318v 00797697v 02502536v 02449847v 00802946v 00803325v 00668099v 00748803v 00748972v 02481436v 02453692v 02481819v 00797697v 01629403v 06689297n"*.

### 4.2. Model training and hyperparameter details

The model is trained on the distributed sense-tagged corpus discussed in Section 4.1. The model was trained using the skip-gram architecture of the Word2Vec model [38] with hierarchical softmax [39] as implemented in GENSIM [53]. The other hyperparameters used for each of the experimental tasks are presented in Table 2. The difference in values of some of these other hyperparameters is because WSD and Word Similarity are two distinct tasks which require different optimal values. The hyperparameters with different values are the embeddings dimension, windows size and subsampling rate. Larger embeddings dimension and window size favors WSD since it allows the model to capture more topical domain information about a word sense while smaller dimension and window size favors Word Similarity since it allows the model to capture more narrow domain information about a word sense which is suitable for computing similarity with other senses. In the case of subsampling rate, lower subsampling rate allows the model to filter out frequently occurring word senses. This is suitable for WSD since these frequently occurring senses constitute noise for accurate sense inference. The reverse is the case for Word Similarity in which higher subsampling rate is favored for computing similarity among senses.

**Table 2**
Other task-specific hyperparameters

| Task | Dimension | Windows size | Minimum count | Epochs | Subsampling |
|---|---|---|---|---|---|
| WSD | 200 | 40 | 1 | 100 | 1e-05 |
| Word Similarity | 100 | 15 | 1 | 100 | 1e-03 |

## 5. Experiments

We carried out experiments on WSD and Word Similarity tasks. The experimental details of the two tasks are presented in sections 5.3 and 5.4 respectively. We adopt the derived sense-specific M-SE distributional similarity measures for the WSD task and the prior measures to the Word Similarity task.

### 5.1. Overview of distributional similarity measures for multi-sense embeddings

Due to the fact that M-SE deal with distinct representation of word senses, it is needful to unify these representations when utilizing them for distributional semantics in natural language processing. As a result of this, a number of measures have been proposed and used in the literature. The first four were proposed by [55]. They are: AvgSim, MaxSim, AvgSimC and MaxSimC. The other two; localSim and globalSim were later proposed. The highlights of the measures are as follow:

- **AvgSim metric:** AvgSim is a measure for computing average similarity of embeddings among the senses of word pairs, $w$ and $w'$. It is defined by Equation (1):

$$\text{AvgSim}(w, w') = \frac{1}{MM'} \sum_{i=1}^{M} \sum_{j=1}^{M'} d(v(w, i), v(w', j)), \quad (1)$$

where $d(v(w, i), v(w', j))$ is the similarity between embeddings of $i$th sense of word $w$ and $j$th sense of word $w'$ denoted $v(w, i)$ and $v(w', j)$ respectively. $M$ and $M'$ are the number of senses obtainable for words $w$ and $w'$, respectively.

- **MaxSim metric:** MaxSim computes the maximum similarity among the senses of word pairs, $w$ and $w'$. It is defined by Equation (2):

$$\text{MaxSim}(w, w') = \max_{1 \leq i \leq M, 1 \leq j \leq M'} d(v(w, i), v(w', j)), \quad (2)$$

where the parameters remain as in Equation (1).

- **AvgSimC metric:** AvgSimC is a measure which computes the average similarities among all the senses of a target word $w$ and the context $w'$ in which the target word appears. The formulation of AvgSimC is given in Equation (3):

$$\text{AvgSimC}(w, w') = \frac{1}{MM'} \sum_{i=1}^{M} \sum_{j=1}^{M'} P(w, c, i) \times \\ P(w', c', j) \times d(v(w, i), v(w', j)), \quad (3)$$

where given the context $C > 1$, $v(w)$ is the embeddings of the $i$th sense of word $w$. $P(w, c, i)$ is the similarity of the $i$th sense of word $w$ with its context $c$. As a result of the peculiarity of multi-sense embeddings which produce embeddings for each sense of a word, Ruas et al. [59] determines AvgSimC$(w, w')$ by averaging embeddings of all senses of context words $w'$, then computes the average of similarities of senses of word $w$ against the averaged embeddings.

- **MaxSimC metric:** MaxSimC computes the maximum similarity among all the senses of the context words $w'$ in which a target word $w$ appears. It is defined by Equation (4):

$$\text{MaxSimC}(w, w') = d(v_{sk}(w, i), v_{sk}(w', j)), \quad (4)$$

where given the context $C > 1$, $v_{sk}(w, i)$ is the maximum similarity among all word senses of word of $w$ with the senses of context words $w'$. Ruas et al. [59] determines MaxSimC$(w, w')$ in similar fashion to AvgSimC$(w, w')$ by computing the maximum similarity among senses of $w$ with the averaged embeddings of context words $w'$.

- **localSim metric:** localSim was introduced by Neelakantan et al. [46] for the representation of each word sense within a local context of usage.

- **globalSim metric:** globalSim was also introduced by Neelakantan et al. [46] and it measures similarity using

each word's global context vector, ignoring their senses. It is defined by Equation (5):

$$\text{globalSim}(w, w') = d(v_g(w), v_g(w')), \quad (5)$$

where $w$ and $w'$ are target and context words, respectively. Ruas et al. [59] also modified globalSim which resulted in a variant that considers senses of the target and context words. This was done by averaging the vectors of senses of target and context words $w$ and $w'$ respectively. $d(v_g(w))$ represents averaged embeddings of all senses of word $w$, while $v_g(w')$ represents averaged embeddings of all word senses of context words $w'$. Therefore, the global similarity is given by the similarity between the averaged embeddings.

### 5.2. Measures for sense-specific distributional semantic similarity

Prior distributional similarity measures for M-SE were tailored towards Word Similarity task and considered only word pairs and their possible context. We introduce sense-specific similarity measures, which compute similarities between each specific senses of word pairs or target word with multiple context words. The basic sense-specific measures are AvgSimS and MaxSimS, from which SumAvgSimS and SumMaxSimS are then derived respectively. The details are as follow:

- **AvgSimS Metric:** AvgSimS computes the average similarity between a specific sense $s_i$ of a target word and the sense(s) of a paired word $w'$. It is defined by Equation 6:

$$\text{AvgSimS}(s_i, w') = \frac{1}{M'} \sum_{j=1}^{M'} d(v(s_i), v(w', j)), \quad (6)$$

where $d(v(s_i), v(w', j))$ is the similarity between embeddings of a specific sense $s_i$ of a target word $w$ and the $j$th sense of a paired word $w'$, respectively. The embeddings are denoted $v(s_i)$ and $v(w', j)$, respectively. $M'$ denotes the number of senses of the context word.

- **SumAvgSimS Metric:** SumAvgSimS computes the cumulative average similarities between a specific sense $s_i$ of a target word and the sense(s) of each context word $w'$ when there are multiple context words where the target word appears, as applicable in most sentences. The cumulative average similarity is given by Equation 7:

$$\text{SumAvgSimS}(s_i, w') = \sum_{k=2}^{N} \left( \frac{1}{M'} \sum_{j=1}^{M'} \left( d(v(s_i), v(w', j)), k \right) \right), \quad (7)$$

where $N$ is the number of context words $w'$ and $\left( d(v(s_i), v(w', j)), k \right)$ is the similarity of $s_i$ and the $j$th sense of $k$th context word $w'$.

- **MaxSimS Metric:** MaxSimS computes the maximum similarity between a specific sense $s$ of a target word and the sense(s) of a paired word $w'$. MaxSimS is defined by Equation 8:

$$\text{MaxSimS}(s_i, w') = \max_{1 \leq j \leq M'} d(v(s_i), v(w', j)), \quad (8)$$

where $d(v(s_i), v(w', j))$ is the similarity between embeddings of a specific sense $s_i$ of a target word $w$ and the $j$th sense of a paired word $w'$. The embeddings are denoted $v(s_i)$ and $v(w', j)$, respectively. $M'$ denotes the number of senses of the context word.

- **SumMaxSimS Metric:** SumMaxSimS computes the cumulative maximum similarity between a specific sense $s$ of a target word and the sense(s) of each context word $w'$ when there are multiple context words where the target word is used, as applicable in most sentences. The cumulative maximum similarity is given by Equation 9:

$$\text{SumMaxSimS}(s_i, w') = \sum_{k=2}^{N} \left( \max_{1 \leq j} \left( d(v(s_i), v(w', j)), k \right) \right), \quad (9)$$

where $N$ is the number of context words $w'$ and $\left( d \left( v(s_i), v(w', j) \right), k \right)$ is the similarity of $s_i$ with the $j$th sense of $k$th context word $w'$.

All similarities used in our experiments are computed using the cosine similarity.

### 5.3. Word sense disambiguation experiment

WSD involves the computational determination of specific senses of words based on their usage in a piece of text. We adapt the sense-specific similarity measures described in Section 5.2 to WSD. Given a target word $w$ with context word(s) $w'_j$, we compute either AvgSimS, MaxSimS, SumAvgSimS, or SumMaxSimS among their senses that are looked up in WordNet. AvgSimS or MaxSimS is used in case of a single-word sentential context while either SumAvgSimS or SumMaxSimS is applied to sentential contexts with multiple words. However, for each experiment, similarity was computed using a measure of a homogeneous concept i.e. either AvgSimS is used jointly with SumAvgSimS or MaxSimS is used jointly with SumMaxSimS. For a target word, the disambiguation algorithm selects its sense with the maximum cumulative similarity score as the appropriate sense for the word. The cumulative similarity score is obtained computing either AvgSimS/SumAvgSimS or MaxSimS/SumMaxSimS depending on the similarity concept adopted.

Algorithm 2 presents the summary of the WSD algorithm. In a situation where a target word lacks context or where there is no similarity, the traditional backoff strategy, a technique which selects a baseline sense of the word as the appropriate sense is used as the backoff though rarely needed except a target word does not have context word. In our case, WordNet's

**Algorithm 2** Word sense disambiguation with multi-sense embeddings
**Require**: Multi-sense embeddings $V$, Lexical Database $D$
**Input**: Text $T = \{w_i, \ldots, w_n\} \in D$
**Output**: Sense $s_i \in D$

1: **procedure** DISAMBIGUATE($w_i$)
2:   **for** $w_i, \ldots, w_n \in T$ **do**
3:     Select $w_t \leftarrow$ target word
4:     Select $w'_j \leftarrow$ context word(s)
5:     Lookup $w_t, w'_j \in D$
6:     **for each** $s_i, \ldots, s_n \in w_t$ **do**
7:       Lookup embeddings $v(s_i) \in V$
8:       Select $s_f \leftarrow$ WordNet 1st sense (WN 1ST)
9:       **if** $w'_j = 1$ **then**
10:         Compute AvgSimS($s_i, w'_j$) | MaxSimS($s_i, w'_j$)
11:         **if** AvgSimS($s_i, w'_j$) | MaxSimS($s_i, w'_j$) $> 0$ **then**
12:           Return $\arg\max_{s_i}$ AvgSimS($s_i, w'_j$) | $\arg\max_{s_i}$ MaxSimS($s_i, w'_j$)
13:         **else**
14:           Return $s_f$
15:         **end if**
16:       **else if** $w'_j > 1$ **then**
17:         Compute SumAvgSimS($s_i, w'_j$) | SumMaxSimS($s_i, w'_j$)
18:         **if** SumAvgSimS($s_i, w'_j$) | SumMaxSimS($s_i, w'_j$) $> 0$ **then**
19:           Return $\arg\max_{s_i}$ SumAvgSimS($s_i, w'_j$) | $\arg\max_{s_i}$ SumMaxSimS($s_i, w'_j$)
20:         **else**
21:           Return $s_f$
22:         **end if**
23:       **else**
24:         Return $s_f$
25:       **end if**
26:     **end for**
27:   **end for**
28: **end procedure**

first sense (WN 1ST) is used as the backoff. However, if a target word lacks content contextual word, it is possible to apply a generative topic model on the corpus to generate contextual information as applicable in [4] though at a computational cost. We did not apply this technique in our approach.

**Evaluation of WSD experiment:** The accuracy of a WSD system depends on the approach used to develop it. Approaches to WSD are categorized into 4, namely: supervised, unsupervised, semi-supervised and knowledge-based systems.

Supervised systems makes use of classifiers which are trained on manually sense-annotated corpora. They are known for their superior performance over other approaches, however, their accuracy depends on the training/examples and require repetition for different test datasets. Furthermore, manual tagging of corpora for the purpose of supervised WSD is a costly and time-consuming process, which makes supervised systems almost impractical in very large corpora. This problem is commonly referred to as the *"knowledge acquisition bottleneck"*.

Unsupervised systems rely on unannotated corpora and induce word senses based on some similarity in contextual information among words. Semi-supervised systems on the other hand, attempt to resolve the *"knowledge acquisition bottleneck"* inherent in supervised systems by using a small tagged corpus as a seed to a bootstrapping process in order to generate a larger annotated corpus.

Finally, knowledge-based systems, also referred to as dictionary-based systems, make use of knowledge in lexical resources such as dictionaries, thesauri, and ontologies. Some of the techniques in this approach are based on the idea that senses can be inferred from the computation of similarity between features of a target word and its context. The foundation of the hypothesis behind these similarity techniques is the *"gloss overlap"* proposed by Lesk [31], which computes similarity based on the overlap of words in the definition of a target word and that of its context words. Other knowledge-based techniques include those based on graphs among others.

Our approach is knowledge-based and relies on the sense embeddings of both target and context words. We evaluate the M-SE-based WSD algorithm on six benchmark datasets. Each of the original datasets were annotated with different versions of WordNet but were later standardized and unified by Raganato et al. [52] with the exception of SemEval-2007 task 07. All the datasets in the unified version and the coarse-grained task are tagged with WordNet 3.0. The descriptions of the datasets are as follow:

- **Senseval-2 (SE2):** The Senseval-2 dataset [16], after standardization, consists of 2,282 sense tags comprising nouns, verbs, adjectives and adverbs. It was originally tagged with WordNet version 1.7 senses.

- **Senseval-3 English all-words task (SE3):** The Senseval-3 English all-words task [62] consists of 1,850 sense annotations. Before standardization, it was annotated with WordNet version 1.7.1.

- **SemEval-2007 task 17 (SE07-17):** The English lexical sample, semantic role labeling SRL and all-words task [51] consists of 455 annotations of only nouns and verbs. Before standardization, it was annotated with WordNet version 2.1.

- **SemEval-2007 task 07 (SE07-07):** The coarse-grained English all-words task [44] consists of 2,269 annotated words from five documents covering the domains of journalism, book review, travel, computer science and biography. The tagged words consist of 1,108 nouns, 591 verbs, 362 adjectives and 208 adverbs. It was also originally annotated with WordNet version 2.1 senses.

- **SemEval-2013 task 12 (SE13-12):** SemEval-2013 task 12 [43] is a multilingual WSD task with the original dataset tagged mainly in BabelNet 1.1.1 with a byproduct in WordNet 3.0 and the Wikipedia sense inventory. Since it is a multilingual task, it also contains subtasks in English, French, Spanish, German, and Italian. The English subtask, which was used for this experiment, consists of 1,644 sense annotations.

- **SemEval-2015 task 13 (SE15-13):** SemEval-2015 task 13 [41] is also a multilingual dataset which combines WSD with entity linking. It consists of 1,022 annotations and covers the domains of biomedicine, computing & mathematics and social issues.

Methods in the different approaches to WSD have diverse underlying ideas and hypotheses. Ours is based on enhanced distributional semantics provided by multi-sense embeddings. Most importantly, our major aim is to evaluate our model for enhanced distributional semantic structure and compared with the primary baselines, which is the model without enhancement. Furthermore, as there are several methods within each of the approaches to WSD, our secondary baselines are the methods based on word/sense embeddings evaluation. However, works on embeddings evaluation for WSD are still very limited.

Table 3 compares WSD evaluation results on our enhanced model (EDS-MEMBED) and the baseline (MEMBED) model with other state-of-the-art embeddings-based systems. It consists of the six major benchmark datasets SE2, SE3, SE07-17, SE13-12, SE15-13 and SE07-07. However, as reported earlier, all the systems evaluates on a subset of the datasets.

A brief highlights of the state-of-the-art embeddings-based systems compared with our systems are given as follow:

- **IMS+Word2vec:** IMS+Word2vec [29] uses Word2Vec embeddings as features to a previous state-of-the-art supervised WSD algorithm called IMS [68]

- **VecLesk(baroni c) & VecLesk(glove):** They are variants of a knowledge-based algorithm [63] which respectively uses *baroni c* and *glove* embeddings for summing the vectors of the terms inside a sense's definition, weighted in function of their part of speech and their frequency

- **+LS:** +LS [48] is an adaptation of the Lesk algorithm that uses word and sense embeddings to compute the similarity between the gloss of a sense and the context of the word

- **Chen et al.(Model S2C):** Chen et al.(Model S2C) [13] presents an approach that is premised on two ideas: (i) that a quality representation of word senses should capture rich information about words and senses, which in turn should be helpful for WSD and (ii) that WSD can provide well-grounded disambiguated corpus for quality sense representations. It evaluates on both knowledge-based and semi-supervised approaches, the results of which are categorized accordingly in the table

- **IMS + S-product:** IMS + S-product [58] uses sense embeddings as features to a supervised algorithm.

Table 4 shows the results of M-SE-based WSD using EDS-MEMBED (our enhanced distributional semantic model) juxtaposed with MEMBED (primary baseline model without enhancement). In addition, the table also compares state-of-the-art knowledge-based and supervised systems.

For the purpose of classification, we generalize MaxSimS and SumMaxSimS as MaxS, since they are used jointly for each experiment as a homogeneous concept. Similarly, AvgSimS and SumAvgSimS using average concept of similarity are generalized as AvgS. EDS-MEMBED$_{\text{MaxS}}$ and EDS-MEMBED$_{\text{AvgS}}$ are the results of our algorithm. In the two cases, the subscript indicates the similarity concepts of maximum and average measures.

Furthermore, some systems evaluated on only a subset of the datasets. The merger of SE2, SE3, SE07-17, SE13-12 and SE15-13, is labeled as "ALL" in the table, and the merger also evaluated at POS levels indicated as Nouns, Verbs, Adj. (for adjectives) and Adv. (for adverbs). Any system with empty column in any of the five datasets belongs to the category which evaluated on only a subset of the datasets. For systems which present multiple runs or variations of their algorithm, only the run with the best overall performance is included for comparison.

A brief highlights of the state-of-the-art systems compared with our system are given as follow:

- **LMMS$_{2348}$(BERT):** LMMS$_{2348}$(BERT) [34] uses contextual embeddings built from some contextual language models as input to a WSD based on K-Nearest

**Table 3**
$F_1$ (%) of EDS-MEMBED in comparison with MEMBED and state-of-the-art embeddings-based WSD systems on the benchmark datasets. Best results for each approach in bold.

| Approach | System | SE2 | SE3 | SE07-17 | SE13-12 | SE15-13 | SE07-07 |
|---|---|---|---|---|---|---|---|
| | IMS+Word2Vec(OMSTI) | **68.3** | 68.2 | **59.1** | - | - | - |
| Supervised | IMS + S-product | 66.8 | **73.6** | - | - | - | - |
| Semi-Supervised | Chen et al.(Model S2C) | - | - | - | - | - | **82.6** |
| | +LS | 58.4 | 59.4 | - | - | - | - |
| | Chen et al.(Model S2C) | - | - | - | - | - | 75.8 |
| Knowledge-based | VecLesk(baroni c) | - | - | - | - | 58.0 | 75.3 |
| | VecLesk(glove) | - | - | - | - | 59.0 | 73.0 |
| | MEMBED$_{AvgS}$ | 65.7 | 58.8 | 47.5 | 64.2 | 69.2 | 78.6 |
| | MEMBED$_{MaxS}$ | 66.1 | 58.9 | **48.4** | 64.4 | 70.3 | 78.9 |
| | EDS-MEMBED$_{AvgS}$ | 66.3 | **60.4** | 46.2 | 65.6 | **72.4** | 78.5 |
| | EDS-MEMBED$_{MaxS}$ | **66.6** | 59.9 | 47.0 | **66.7** | 71.9 | **79.1** |

**Table 4**
$F_1$ (%) of EDS-MEMBED in comparison with MEMBED and state-of-the-art WSD systems on the benchmark datasets. Best results for each approach in bold. † indicates values not given by authors but deduced [65].

| Approach | System | SE2 | SE3 | SE07-17 | SE13-12 | SE15-13 | Nouns | Verbs | Adj. | Adv. | ALL | SE07-07 |
|---|---|---|---|---|---|---|---|---|---|---|---|---|
| | LMMS$_{2348}$(BERT) | 76.3 | 75.6 | 68.1 | 75.1 | 77.0 | - | - | - | - | 75.4 | - |
| | HCAN | 72.8 | 70.3 | - | 68.5 | 72.8 | 72.7 | 58.2 | 77.4 | 84.1 | 71.1 | - |
| | LSTMLP | 73.8 | 71.8 | 63.5 | 69.5 | 72.6 | †73.9 | - | - | - | †71.5 | 83.6 |
| | GAS | 72.2 | 70.5 | - | 67.2 | 72.6 | 72.2 | 57.7 | 76.6 | 85.0 | 70.6 | - |
| Supervised | BERT | 75.5 | 73.6 | 68.1 | 71.1 | 76.2 | - | - | - | - | 74.1 | - |
| | GLOSSBERT | 77.7 | 75.2 | 72.5 | 76.1 | 80.4 | 79.3 | 66.9 | 78.2 | 86.4 | 77.0 | - |
| | SemCor+WNGC | **79.7** | **77.8** | **73.4** | **78.7** | **82.6** | **81.4** | **68.7** | **83.7** | **85.5** | **79.0** | **90.4** |
| | MFS | 65.6 | 66.0 | 54.5 | 63.8 | 67.1 | 67.7 | 49.8 | 73.1 | 80.5 | 65.5 | 78.9 |
| | KEF | **69.6** | 66.1 | **56.9** | 68.4 | 72.3 | **71.9** | **51.6** | 74.0 | 80.6 | **68.0** | - |
| | Babelfy | 67.0 | 63.5 | 51.6 | 66.4 | 70.3 | 68.6 | 49.9 | 73.2 | 79.8 | 65.5 | - |
| | UKB | 68.8 | 66.1 | 53.0 | **68.8** | 70.3 | - | - | - | - | 67.3 | - |
| | WSD-TM | 69.0 | **66.9** | 55.6 | 65.3 | 69.6 | 69.7 | 51.2 | **76.0** | 80.9 | 66.9 | - |
| Knowledge-based | WN 1st sense | 66.8 | 66.2 | 55.2 | 63.0 | 67.8 | 67.6 | 50.3 | 74.3 | 80.9 | 65.2 | 78.9 |
| | MEMBED$_{AvgS}$ | 65.7 | 58.8 | 47.5 | 64.2 | 69.2 | 66.7 | 47.2 | 69.3 | 73.7 | 62.9 | 78.6 |
| | MEMBED$_{MaxS}$ | 66.1 | 58.9 | 48.4 | 64.4 | 70.3 | 66.8 | 48.7 | 68.8 | 75.7 | 63.3 | 78.9 |
| | EDS-MEMBED$_{AvgS}$ | 66.3 | 60.4 | 46.2 | 65.6 | **72.4** | 68.2 | 48.4 | 69.9 | 74.9 | 64.2 | 78.5 |
| | EDS-MEMBED$_{MaxS}$ | 66.6 | 59.9 | 47.0 | 66.7 | 71.9 | 68.0 | 49.2 | 70.3 | 76.0 | 64.4 | **79.1** |

Neighbors (K-NN). They experimented with diverse language models but the one based on BERT [14] is the best performing.

- **HCAN**: HCAN [35] leverages the glosses of word senses in neural WSD using co-attention technique.

- **LSTMLP**: LSTMLP [67] uses long short-term memory (LSTM) to capture sequential and syntactic patterns in text for WSD.

- **GAS**: GAS [36], similar to Luo et al. [35] incorporates word sense glosses, jointly with context into the the training of a neural WSD algorithm.

- **BERT**: BERT [21] employs contextualized word repres-

entations for WSD and produces two variations of the algorithm.

- **GLOSSBERT**: GLOSSBERT [26] leverages and incorporates glosses of word senses in a lexical resource into a contextualized neural WSD.

- **SemCor+WNGC**: SemCor+WNGC [65] relies on relations in WordNet for vocabulary compression in neural WSD.

- **KEF**: KEF [66] models WSD as semantic space and semantic path by Latent Semantic Analysis (LSA) and PageRank, respectively, using WordNet as the knowledge resource.

- **Babelfy**: Babelfy [42] is a unified graph-based approach to entity linking and WSD based on a loose identification of candidate meanings and subgraph heuristics.

- **UKB**: UKB [1, 2] is a graph-based technique for WSD which applies PageRank and Personalized PageRank on the relations in a lexical knowledge-base.

- **WSD-TM**: WSD-TM [11] develops a system with a variant of the Latent Dirichlet Allocation (LDA) model [6], which utilizes the whole words in a document against the traditional sentential context by replacing topic proportions in a document with synset proportions.

Most frequent sense (MFS) and WordNet first sense (WN 1ST SENSE) are the baselines for supervised and knowledge-based approaches, respectively.

### 5.4. Word similarity experiment

The Word Similarity task can either be one in which word pairs are compared or one with additional context apart from the pair. We applied the AvgSim, MaxSim, and globalSim measures. Word senses of the target words and their context are looked up in WordNet. The embeddings of each word sense are the averaged embeddings of itself and that of its relations in WordNet. The relations used for the representation include hypernymy, hyponymy, causes, entailment, topic domain, pertainymy and derivationally related forms. The summary of the Word Similarity algorithm is presented in Algorithm 3.

**Evaluation of word similarity experiment:** We evaluate our multi-sense embeddings on five benchmark datasets. The description of these datasets are as follow:

- **RG65**: RG65 [60] is one of the oldest Word Similarity datasets. It consists of 65 noun pairs with similarity ranging from 0 to 4.

- **MC28**: MC28 [56] comprises 28 noun pairs and the similarity ranges from 0 to 4.

- **MEN**: MEN [9] consists of 2,000 words pairs in development and 1,000 pairs in test part making a total of 3,000 word pairs. Similarity ranges from 0 to 50.

- **SimLex999**: SimLex999 [24] consists of 999 word pairs and similarity ranges from 0 to 10. The breakdown of the words are 666 noun-noun, 222 verb-verb and 111 adjective-adjective pairs.

- **WordSim-353**: WordSim-353 [20] consists of 353 word pairs divided into two sets of 153 and 200 words, respectively. Similarity ranges from 0 to 10.

Ruas et al. [59] normalized the similarity range of all the datasets to −1 and 1 which is used in this study.

Tables 5 and 6 show the result of our enhanced model (EDS-MEMBED) compared to MEMBED (our baseline model without enhancement) and state-of-the-art embeddings on Word Similarity tasks in terms of the Spearman's correlation coefficient ($\rho$). For models with corpus or algorithm variants or in which the algorithm was trained on multiple corpora, only the variant on any corpus which performs best across the similarity concepts are included for comparison. Systems which do not carry out evaluation on a particular dataset are marked "-" in the result column. The compared systems are briefly described below:

- **GloVe**: GloVe [49] is a distributed word representation technique which considers a word's global occurrence along with its local co-occurrences. It was trained several times on corpora of different sizes, the largest containing 42 billion tokens.

- **Retro**: Retro [17] is a method for retrofitting word vectors through the relations in lexicons in such a way that linked words produce similar vectors.

- **DeConf**: [50] takes pre-trained word or sense embeddings as input and uses a graph-based technique for extracting semantic biasing words from their synsets in a lexicon.

- **Pruned-TF-IDF**: Pruned-TF-IDF [54] is based on the concept of *tiered clustering* for distributed representation of word meaning.

Other systems which include *Word2Vec* [38], SENSEMBED [28], MSSA [59], Huang et al. [25], MSSG [46], SW2V [37], and the system of Chen et al. [13] have all been discussed in detail in Section 2.

### 5.5. Ablation study on the optimality of task-specific hyperparameters

In Tables 7 and 8, we show the result of an ablation study which exchanges the hyperparameter values for the WSD with the optimal values for Word Similarity task, and vice-versa respectively. This is to show the optimality of the task-specific hyperparameter values that we chose and used as discussed in Section 4.2. For convenience, we denote hyperparameter values as HV in the tables.

**Table 5**
Spearman's correlation coefficient ($\rho \times 100$) of models on the RG65, MC28, and MEN datasets. Best performing models in bold.

|   | **RG65** | | | **MC28** | | | **MEN** | | |
| --- | --- | --- | --- | --- | --- | --- | --- | --- | --- |
| **Model** | AvgSim | MaxSim | globalSim | AvgSim | MaxSim | globalSim | AvgSim | MaxSim | globalSim |
| GloVe | - | - | 82.9 | - | - | 83.6 | - | - | - |
| Retro | - | - | 84.2 | - | - | - | - | - | 75.9 |
| Word2Vec | - | - | 75.4 | - | - | - | - | 75.0 | - |
| SENSEMBED | 87.1 | 89.4 | - | - | 88.0 | - | **80.5** | 77.9 | - |
| MSSA | 82.8 | 87.8 | 85.9 | 84.5 | 88.8 | 87.5 | 78.5 | 74.4 | 79.5 |
| DeConf | - | **89.6** | - | - | - | - | - | 78.6 | - |
| SW2V | - | 74.0 | - | - | - | - | - | 76.0 | - |
| **MEMBED** | 64.4 | 85.0 | 83.6 | 70.2 | 86.3 | 81.7 | 55.2 | 67.9 | 58.3 |
| **EDS-MEMBED** | 77.1 | **89.6** | 86.9 | 83.2 | **89.6** | 88.4 | 63.3 | 68.4 | 64.3 |

**Table 6**
Spearman's correlation coefficient ($\rho \times 100$) of models on SimLex999 and WordSim-353. Best performing models in bold.

|   | **SimLex999** | | | **WordSim-353** | | |
| --- | --- | --- | --- | --- | --- | --- |
| **Model** | AvgSim | MaxSim | globalSim | AvgSim | MaxSim | globalSim |
| GloVe | - | - | - | - | - | 75.9 |
| Retro | - | - | - | - | - | 61.2 |
| Word2Vec | - | 39.0 | - | - | - | - |
| SENSEMBED | - | - | - | **77.9** | 71.4 | - |
| Huang et al. | - | - | - | 64.2 | - | 22.8 |
| Pruned-TF-IDF | - | - | - | - | - | 73.4 |
| MSSG | - | - | - | 68.6 | - | 69.1 |
| MSSA | 46.9 | 38.5 | 43.9 | 73.0 | 66.2 | 73.7 |
| DeConf | - | 51.7 | - | - | - | - |
| SW2V | - | 47.0 | - | - | 71.0 | - |
| Chen et al. | - | 43.0 | - | - | - | - |
| **MEMBED** | 37.0 | 47.6 | 33.2 | 45.6 | 54.0 | 44.6 |
| **EDS-MEMBED** | 48.0 | **53.7** | 46.4 | 59.5 | 59.6 | 58.2 |

**Table 7**
$F_1$ (%) of EDS-MEMBED on WSD tasks using the optimal and exchange hyperparameter values. Best results in bold

|   | **Optimal HV** | | **Exchange HV** | |
| --- | --- | --- | --- | --- |
| **WSD task** | EDS-MEMBED$_{\text{AvgS}}$ | EDS-MEMBED$_{\text{MaxS}}$ | EDS-MEMBED$_{\text{AvgS}}$ | EDS-MEMBED$_{\text{MaxS}}$ |
| SE2 | 66.3 | **66.6** | 61.0 | 62.1 |
| SE3 | **60.4** | 59.9 | 56.2 | 57.4 |
| SE07-17 | 46.2 | **47.0** | 41.5 | 42.9 |
| SE13-12 | 65.6 | **66.7** | 59.8 | 60.8 |
| SE15-13 | **72.4** | 71.9 | 65.9 | 66.7 |
| SE07-07 | 78.5 | **79.1** | 75.7 | 75.5 |

**Algorithm 3** Word similarity with multi-sense embeddings

**Require**: Multi-sense embeddings $V$, Lexical Database $D$
**Input**: Words $T = \{w_i, \ldots, w_n\} \in D$
**Output**: Similarity Score, $Sc$

```
 1: procedure SIMILARITY(w_i, w'_j)
 2:     for w_i, …, w_n ∈ W do
 3:         Select w_t ← target word
 4:         Select w'_j ← context word(s)
 5:         Lookup w_t, w'_j ∈ D
 6:         s_i, …, s_n ← senses of w_t and w'_j
 7:         Lookup embeddings v(s_i), …, v(s_n) ∈ V and compute the averaged embeddings from sense relations
 8:         if w'_j = 1 then
 9:             Compute AvgSim(w_t, w'_j) | MaxSim(w_t, w'_j) | globalSim(w_t, w'_j)
10:             Return Sc = AvgSim(w_t, w'_j) | MaxSim(w_t, w'_j) | globalSim(w_t, w'_j)
11:         else if w'_j > 1 then
12:             Compute AvgSimC(w_t, w'_j) | MaxSimC(w_t, w'_j) | globalSim(w_t, w'_j)
13:             Return Sc = AvgSimC(w_t, w'_j) | MaxSimC(w_t, w'_j) | globalSim(w_t, w'_j)
14:         else
15:             Return Sc = 0
16:         end if
17:     end for
18: end procedure
```

**Table 8**

Spearman's correlation coefficient ($\rho \times 100$) of EDS-MEMBED on Word Similarity tasks using the optimal and exchange hyperparameter values. Best results in bold

| WSD task | Optimal HV | | | Exchange HV | | |
| --- | --- | --- | --- | --- | --- | --- |
| | AvgSim | MaxSim | globalSim | AvgSim | MaxSim | globalSim |
| RG65 | 77.1 | **89.6** | 86.9 | 80.6 | 88.2 | 82.6 |
| MC28 | 83.2 | **89.6** | 88.4 | 81.7 | 88.3 | 88.5 |
| MEN | 63.3 | 68.4 | 64.3 | 68.2 | 68.8 | **69.6** |
| SimLex999 | 48.0 | **53.7** | 46.4 | 49.4 | 48.7 | 44.8 |
| WordSim-353 | 59.5 | **59.6** | 58.2 | 59.4 | 58.0 | 58.4 |

### 5.6. Notes on models' computational time

As discussed in Section 1 that one of the advantages of our multi-sense model is the efficiency in training, we juxtapose the sizes of the training data, the computational time, the dimensions and the hardware environment under which they run for all the referenced models in Table 9. Though the time are not specified for some of the models but the size of the training data can provide some insights since the computational time is directly proportional to the data size. The models listed in the evaluation of Word Similarity task but not listed in this table simply means they use one or more of the ones listed here as input to their models.

## 6. Discussion of results

The results show that EDS-MEMBED (enhanced model) improves the quality of embeddings when compared to MEMBED, the model built from the original corpora without the enhancement.

On the WSD experiment, the result on the SE2 task shows that the best of EDS-MEMBED and MEMBED have $F_1$ of 66.6% and 61.1%, respectively. In the SE3 task, the improvement in the $F_1$ value is 1.5% for the best runs of the two models. The enhancement generally improves performance in all the tasks except SE07-17. The improvement is more pronounced in the SE15-13 task with a difference of 2.1% in $F_1$ values between the best of EDS-MEMBED and MEMBED.

The comparison in Table 3 shows MEMBED and EDS-MEMBED outperforms other state-of-the-art embeddings-based methods on the knowledge-based approach across all the datasets. Chen et al.(Model S2C) [13] is the only semi-supervised approach and was evaluated only on SE07-07 dataset with $F_1$ of 82.6%. For supervised systems, IMS+Word2vec [29] achieves the best performance on SE2 and SE07-17 datasets with $F_1$ values of 68.3% and 59.1% respectively. IMS + S-product [58] achieves the best result on SE3 with 73.6% $F_1$ value. However, on the general comparison with methods of diverse ideas, the results present a mixed set of strengths and weaknesses. KEF [66] is the predominantly better system than others in the knowledge-based approach while EDS-MEMBED produces competitive performance with the state-of-the-art knowledge-based algorithms and achieves the best $F_1$ on SE15-13 and SE07-07 datasets. Overall, for knowledge-based systems, KEF [66] achieves state-of-the-art results in most of the datasets. Semcor+WNGC [65] performs best for supervised systems across all the datasets. We deliberately experimented with the SE07-07 dataset which is often ignored by most previous works. This is consistent with our goal of making multi-sense embeddings useful as part of the processing pipelines for other semantic tasks. This is due to the fact that in some of these tasks, a coarse sense of a word is sufficient for representation. For instance, knowing that the verb senses of the word *"buy"* are about the exchange of good or services for money or other medium of exchange is sufficient. WordNet has about three fine-grained senses which can be merged to represent the single idea about exchange of goods and services using an exchange medium.

**Table 9**
Training data size, training time, dimension of the models and the hardware environment

| Model | Data size (tokens) | Training time | Dimension | Hardware |
| --- | --- | --- | --- | --- |
| Huang at al. | 990 million | 168h | 50 | - |
| MSSG | 990 million | 6h | 300 | - |
| SENSEMBED | 3 billion | - | 400 | - |
| MSSA | 540 million | - | 1000 | - |
| Word2Vec | 783 million | 24h | 300 | - |
| GloVe | 840 billion | 85m/6billion | 300 | Intel CPU @ 2.1GHz × 32 |
| MEMBED | 1,302,549 | 12m:19s | 100 | Intel CPU @ 1.60GHz × 4, 11.7 GiB RAM |
| EDS-MEMBED | 1,590,998 | 16m:26s | 200 | Intel CPU @ 1.60GHz × 4, 11.7 GiB RAM |

In the Word Similarity task, the effect of enhancement is markedly better than in WSD task. On the RG65 dataset, Spearman's correlation coefficients of the best of EDS-MEMBED and MEMBED are 89.6% and 85.0%, respectively, with an improvement of 4.6% in correlation. EDS-MEMBED and SENSEMBED are the best performing models on the RG65 dataset with 89.6% correlation. On the MC28 dataset, EDS-MEMBED improves correlation by 3.6% and achieves state-of-the-art performance among all the compared systems with 89.6% correlation. In the case of the MEN dataset, the correlations for the best of EDS-MEMBED and MEMBED are 68.4% and 67.9%, respectively, with an improvement of 0.5%. SENSEMBED is the best performing model on the MEN dataset with correlation of 80.5%. On SIMLEX999, EDS-MEMBED has the best performance of 53.7% correlation with an improvement of 6.1% over the best of MEMBED. The improvement of EDS-MEMBED over MEMBED in their best results on the WORDSIM-353 dataset is 5.6% with 59.6% and 54.0% correlations, respectively. The best performing system on this dataset is the SENSEMBED [28] with 77.9% correlation. Considering the performance of the three similarity concepts, the maximum similarity concept tend to generally outperforms the concept of average similarity. The average similarity concept in turn performs better than global similarity concept.

Most of the existing embeddings-based methods for both WSD and Word Similarity evaluates on selected subset of the datasets, probably the ones on which they perform best. We have carried out a robust evaluation across all the datasets to provide insight on the peculiarity of each of the datasets which may trigger future investigation. The only logical conclusion that can be made on why EDS-MEMBED does not achieve superior performance on some datasets on both tasks can be attributed to lack of adequate representation for the word senses which composed the datasets. In view of this, as discussed in Section 4.1, the performance of multi-sense embeddings is largely dependent on the size of the training data and the accuracy of the annotations. The performance of EDS-MEMBED on WSD and Word Similarity tasks shows that it can serve as a source of high-quality sense embeddings for supervised algorithms in many natural language processing NLP applications.

Tables 7 and 8 show the results of an ablation study on the optimality of the hyperparameter values used for each of the tasks compared to that of their exchange values. The optimal values outperforms the exchange values across almost all the tasks.

On a final note, as shown in Table 9, it takes just twelve minutes, nineteen seconds and sixteen minutes, twenty six seconds to train MEMBED and EDS-MEMBED respectively. This is an extremely computationally efficient when compared to other models even at much lover hardware configuration. Furthermore, a close look at the training data sizes of the models shows that our data is just a minute fraction of the others. Logically speaking, this suggests that if a considerable additional fraction is added for training, it may achieve a superior performance across board. However, as stated earlier, it is largely dependent on the accuracy of the tagged word senses.

## 7. Conclusion

Word embeddings conflate the meanings of words and therefore suffer from the problem of polysemy. Multi-sense embeddings are a proposition to resolve this problem of meaning in distributed word representation. However, even with multi-sense embeddings, learning distributional semantics is still hampered by insufficient distributional semantic structures in the training data. In this paper, we propose an approach that is exhaustive of all possible word senses and we show how the distributional structures in training data can be enhanced by leveraging word sense relations in a lexical knowledge-base from which multi-sense embeddings are built. We experiment with modestly-sized training data and show that our algorithm improves the quality and the effectiveness of multi-sense embeddings. Furthermore, we adapt prior embeddings' similarity measures to a knowledge-based word sense disambiguation (WSD) algorithm. Experimental results on WSD and Word Similarity tasks involving 11 benchmark datasets show that our algorithm achieves state-of-the-art performance in some of the datasets. Training on the distributed corpus and additional larger data with high accuracy of sense annotations can further improve the quality of the model and consequently the performance.

## Acknowledgements

The first author's work was supported by the Faculty of Informatics, Masaryk University. The third author's work was graciously funded by the South Moravian Centre for International Mobility as a part of the Brno Ph.D. Talent project.

## CRediT authorship contribution statement

**Eniafe Festus Ayetiran:** Conceptualization, Methodology, Software, Evaluation, Manuscript. **Petr Sojka:** Methodology, Evaluation, Manuscript, Project Administration. **Vít Novotný:** Software, Evaluation, Manuscript.